\pdfoutput=1

\documentclass[11pt]{article}

\usepackage{EACL2023}

\usepackage{times}
\usepackage{latexsym}

\usepackage[T1]{fontenc}

\usepackage[utf8]{inputenc}

\usepackage{graphicx}
\usepackage{subcaption}
\usepackage{multirow}
\usepackage{booktabs}
\usepackage{enumitem}
\setlist{nolistsep}
\usepackage{xspace}
\usepackage{amsmath}

\newcommand{\roberta}{\textsc{RoBERTa}\xspace}
\newcommand{\electra}{\textsc{ELECTRA}\xspace}

\newcommand{\tfive}{\textsc{T5}\xspace}
\newcommand{\gpt}{\textsc{Gpt-2}\xspace}
\newcommand{\gpts}{\textsc{Gpt2}-small\xspace}
\newcommand{\gptm}{\textsc{Gpt2}-medium\xspace}
\newcommand{\gptl}{\textsc{Gpt2}-large\xspace}

\newcommand{\sql}{\textsc{SQL}\xspace}

\newcommand{\squad}{\textsc{SQuAD}\xspace}

\newcommand{\snli}{\textsc{Snli}\xspace}
\newcommand{\boolq}{\textsc{BoolQ}\xspace}
\newcommand{\spider}{\textsc{Spider}\xspace}

\newcommand{\ignore}[1]{}

\definecolor{cb-blue}{RGB}{ 0, 109, 219}
\newcommand{\cbBlue}[1]{\textcolor{cb-blue}{#1}}

\usepackage{microtype}

\usepackage{inconsolata}

\title{Benchmarking Long-tail Generalization with Likelihood Splits}

\author{Ameya Godbole \and Robin Jia \\
  University of Southern California \\
  \texttt{\{ameyagod,robinjia\}@usc.edu} \\
}

\begin{document}
\maketitle
\begin{abstract}
In order to reliably process natural language, NLP systems must generalize to the long tail of rare utterances. We propose a method to create challenging benchmarks that require generalizing to the tail of the distribution by re-splitting existing datasets. We create `Likelihood Splits' where examples that are assigned lower likelihood by a pre-trained language model (LM) are placed in the test set, and more likely examples are in the training set. This simple approach can be customized to construct meaningful train-test splits for a wide range of tasks. Likelihood Splits surface more challenges than random splits: relative error rates of state-of-the-art models increase by 59\% for semantic parsing on \spider, 93\% for natural language inference on \snli, and 33\% for yes/no question answering on \boolq, on our splits compared with the corresponding random splits. Moreover, Likelihood Splits create fairer benchmarks than adversarial filtering; when the LM used to create the splits is also employed as the task model, our splits do not unfairly penalize the LM.
\end{abstract}

\section{Introduction}
\label{sec:introduction}

\begin{figure}[tb]
\centering
\includegraphics[width=0.49\textwidth]{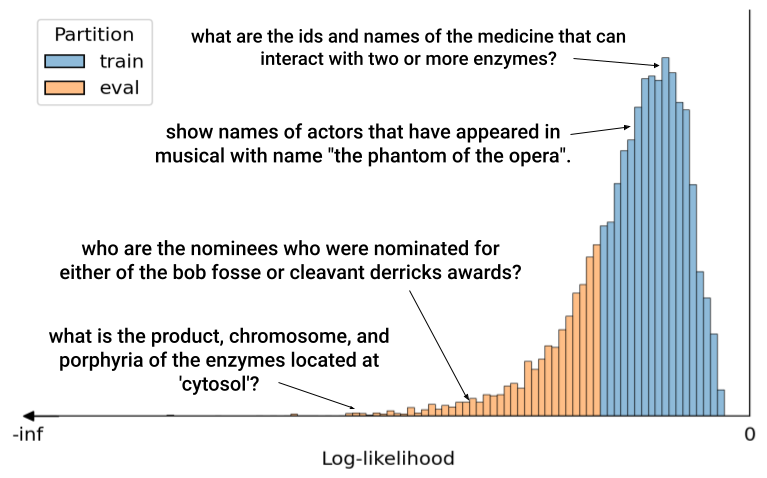}
\caption{\textbf{Likelihood Splits}: We propose to partition the dataset based on likelihood under a language model. The high-likelihood ``head'' of the distribution becomes the training set while we evaluate generalization to the low-likelihood ``tail'' of the data. Shown here are queries from the \spider dataset in different likelihood buckets: one possible tail generalization could be the handling uncommon entities with known query types.}
\label{fig:method-eye-candy}
\end{figure}

Success on in-distribution test data does not necessarily show that a system has solved
the underlying task at hand. Systems can achieve artificially high accuracy by exploiting dataset-specific shortcuts, such as spurious feature-label correlations that hold in the data but not in general \citep{gardner-etal-2021-competency}. 
In many datasets, a large proportion of test examples are similar to training examples, further inflating in-distribution accuracy \citep{lewis-etal-2021-question, czarnowska-etal-2019-dont-forget-tail,bootleg-self-supervision}. Out-of-distribution (OOD) evaluation paints a clearer picture of a system's ability to perform the task.

Prior work has proposed a variety of methods to test OOD generalization, each with their own strengths and weaknesses. Task-specific behavior tests \citep{ribeiro-etal-2020-beyond-checklist,naik-etal-2018-stress,gardner-etal-2020-evaluating-contrast-sets} give insights into model behavior but require significant manual (often expert) effort to create. Adversarial data collection, in which annotators try to fool high-performing models \citep{nie-etal-2020-adversarial-anli,potts-etal-2021-dynasent}, also collects challenging examples, but runs the risk of focusing only on a narrow subset of model weaknesses \citep{bowman-dahl-2021-will,kaushik-etal-2021-efficacy}.
Adversarial filtering removes easy examples from existing datasets \citep{10.1145/3474381-winogrande}, but can disproportionately penalize the model used during filtering \citep{arxiv.2111.08181-unfair-adversarial}. Domain generalization tests transferability to new data domains \citep{fisch2019mrqa,pmlr-v119-miller20a-squad-ood}, but there is no guarantee that generalizing to a given new domain is possible---out-of-domain examples may require skills that are not learnable from the training data \citep{geiger-etal-2019-posing-fair-generalization-tasks}. Other approaches create dataset splits that test for specific skills, such as length generalization \citep{SCAN-dataset} and compositional generalization \citep{shaw-etal-2021-compositional-tmcd}, but they only apply to a narrow subset of tasks. 

In this work, we propose \textbf{Likelihood Splits}, a general-purpose method to create challenging OOD splits for existing datasets. 
The principle behind Likelihood Splits is that any system that claims to reliably process natural language must be able to generalize from more common utterances seen during training to the long tail of rare utterances at test time.
Generalization, not merely memorization, is necessary because even a very large training dataset cannot exhaustively cover all possible long-tail examples that may be encountered in the real world.
Moreover, standard annotation procedures tend to over-sample examples from the head of the distribution, further ignoring the challenge posed by infrequent examples. We identify tail examples using the likelihood under the \gpt language model \citep{radford2019language-gpt2}. Examples with low likelihood under \gpt are placed in the held-out evaluation sets and the high likelihood examples are used as the training set (see Figure~\ref{fig:method-eye-candy}).

Likelihood Splits are a novel, widely applicable strategy that can create interesting generalization benchmarks at no additional annotation cost.
They are more challenging than a random split across a wide range of tasks: error rates relative to random splits increase by 59\% for \tfive \citep{raffel2020-t5} on \spider \citep{yu-etal-2018-spider}, 93\% for \electra \citep{clark2020electra} on \snli \citep{bowman-etal-2015-large-snli}, and 33\% for \roberta \citep{liu2019roberta} on \boolq \citep{clark-etal-2019-boolq}. 
Moreover, the proposed splits do not unfairly penalize the \gpt model used to create the splits when it is used as a task model, thus avoiding one of the downsides of adversarial filtering.
We identify many independent challenges required by Likelihood Splits, 
including generalizing to rare words, complex programs, and syntactically complex sentences.
We encourage future benchmark creators to release Likelihood Splits as a complementary evaluation to the standard IID evaluation to better test out-of-distribution generalization performance. We will release the splits discussed in this work along with the code to easily create Likelihood Splits of other datasets.\footnote{\href{https://github.com/ameyagodbole/long-tail-likelihood-splits}{github.com/ameyagodbole/long-tail-likelihood-splits}}

\section{Related Work}
\label{sec:related-work}




\paragraph{Generalizing to the long-tail.}
Evaluating systems on long-tail phenomena is important, especially because many datasets over-sample the head of the distribution. For example, some question-answering (QA) datasets limit their purview to popular web-pages \citep{yang2018hotpotqa} or frequent user queries \citep{47761-natural-questions}. 
\citet{lewis-etal-2021-question,DBLP:journals/corr/abs-2109-01156} demonstrate that models trained on these datasets often fail on examples that do not match the most frequent training cases. Similar observations have been made in entity linking to rare entities, \citep{bootleg-self-supervision,chen-etal-2021-evaluating-amber-ir}, information retrieval for open-domain QA \citep{sciavolino-etal-2021-simple}, relation extraction for rare relations \citep{sabo-etal-2021-revisiting} and lexicon induction for rare senses in machine translation \citep{czarnowska-etal-2019-dont-forget-tail}. Zero-shot performance of large LMs on numerical reasoning and factoid questions is also correlated with the frequency of occurence of the facts in the pre-training corpus \citep{https://doi.org/10.48550/arxiv.2202.07206-long-tail-numeracy,https://doi.org/10.48550/arxiv.2211.08411-long-tail-closed-book-qa,https://doi.org/10.48550/arxiv.2207.14251-causal-pretraining-effects}. While we do not test whether models can memorize long-tail knowledge, we instead test whether models can process long-tail sentences. \citet{naik-etal-2022-adapting-macro-long-tail} note that it is challenging to catalogue and evaluate generalization along micro-level dimensions 
and instead propose benchmarks that vary along macro-level dimensions (such as the language and domain) as a proxy.
We hypothesize that LMs learn which micro-level phenomena are rare, as this would improve their overall language modeling objective. In this work, we present a recipe that leverages LMs to evaluate tail generalization for any language task.

\paragraph{Task-specific test sets.}
\citet{ribeiro-etal-2020-beyond-checklist} use templated queries to evaluate model performance under various linguistic perturbations. This method requires dataset designers to define phenomena of interest and axes of perturbation along which labels may be preserved or changed. \citet{naik-etal-2018-stress} analyze model errors and instantiate tests that explicitly evaluate models on more examples from each error class. \citet{gardner-etal-2020-evaluating-contrast-sets} check for model consistency under local perturbations of test set examples. All of these approaches require annotators to create new examples, whereas we propose a method to resplit existing datasets.

\paragraph{Adversarial approaches.}
\citet{sogaard-etal-2021-need-to-talk-about-random-splits} argue that random splits over-estimate model performance on new in-domain data and recommend the use of adversarial and heuristically challenging splits to estimate generalizability. Adversarial data collection promotes the creation of difficult examples by encouraging annotators to fool a model-in-the-loop \citep{nie-etal-2020-adversarial-anli, potts-etal-2021-dynasent,kiela-etal-2021-dynabench}. Similarly, Adversarial Filtering removes examples that are easy for a given task model in order to create more challenging benchmarks \citep{10.1145/3474381-winogrande,yang2018hotpotqa}. However, \citet{kaushik-etal-2021-efficacy} and \citet{bowman-dahl-2021-will} point out that adversarially collected or filtered examples may focus on a narrow set of skills that the ``in-the-loop'' model lacks, instead of covering all the abilities required for the underlying task. Additionally, the ``in-the-loop'' task model is disproportionately penalized by the adversarial test sets \citep{arxiv.2111.08181-unfair-adversarial}. We show in \S\ref{sec:model-performance} that Likelihood Splits do not suffer from this issue.

\paragraph{Domain shift.} In NLP, domains can be characterized by the changes in vocabulary and distribution of word use, styles used by authors, and the intended audience. \citet{fisch2019mrqa} pose the challenge of developing QA systems that need to generalize to unseen domains. \citet{pmlr-v119-miller20a-squad-ood} show that QA models trained on \squad show a performance drop on new domains (while human baseline performance remains unchanged); \citet{pmlr-v139-miller21b,hendrycks-etal-2020-pretrained} inter alia perform similar analyses of domain shift. \spider \citep{yu-etal-2018-spider} and \textsc{GrailQA} \citep{gu2021beyond-grailqa} evaluate semantic parsing on unseen table and knowledge base domains respectively. Domain shift is an orthogonal axis of generalization; we focus on generalizing to rare utterances in the same domain.

\paragraph{Out-of-distribution detection.} Previous work in OOD detection has used high generative model perplexity as a sign of outliers \citep{arora-etal-2021-types,NEURIPS2019_1e795968,NEURIPS2018_abdeb6f5}. Our intuition is similar: low likelihood (high perplexity) is an indicator of rare examples. However, only our work uses likelihood scores for benchmark creation. Moreover, in our setting all examples have been collected under the same data collection protocol, so none of the examples are truly OOD.

\paragraph{Compositional generalization.} The ability to ``compose'' the meaning of a new utterance from the known meaning of its parts \citep{Fodor1988ConnectionismAC} is an important aspect of language understanding. The deterministic grammar of programming languages makes semantic parsing, the task of translating a natural language utterance into a logical program, a good testbed for evaluating compositional generalization \citep{SCAN-dataset,COGS-Dataset,PCFG-Set-Dataset,CFQ-Dataset,shaw-etal-2021-compositional-tmcd}. 
However, for tasks where the constituent blocks are not clearly defined, it is unclear how to create such evaluation splits of the data. We compare against compositional generalization splits of the semantic parsing dataset \textsc{Spider} \citep{yu-etal-2018-spider} in \S\ref{sec:experiments}.

\section{Capturing the Tail of the Distribution}
\label{sec:data-split}


In order to find the tail within a dataset, we approximate likelihood of an utterance in the real distribution with its likelihood under a language model (LM). Our method can be easily modified to create meaningful splits for any language task. We demonstrate this by creating Likelihood Splits for:
\begin{itemize}[noitemsep,topsep=0pt,leftmargin=*]
\item \spider, a semantic parsing dataset \citep{yu-etal-2018-spider} consisting of natural language questions and corresponding \sql programs;
\item \snli, a natural language inference dataset \citep{bowman-etal-2015-large-snli} consisting of premise and hypothesis sentences paired with labels denoting that the hypothesis is entailed by/neutral to/contradictory to the premise;
\item \boolq, a question-answering dataset \citep{clark-etal-2019-boolq} consisting of a passages, associated questions, and binary yes/no labels.
\end{itemize}

\subsection{General Approach}
We consider language tasks where models must map an input $x$ to an output $y$ (e.g., a SQL query or a label). 
The input $x$ may be either a single sentence (e.g., semantic parsing) or a pair of sentences (e.g., natural language inference), in which case we write $x = (x_1, x_2)$.
Given a dataset $D$ of $(x, y)$ pairs and desired proportion $p$ of evaluation examples, our method partitions $D$ into subsets $D_{\text{train}}$ and $D_{\text{eval}}$ where $|D_{\text{eval}}| \approx p \cdot |D|$.
More specifically, we will first assign a likelihood score $s(x)$ to each $x \in D$,
then choose $D_{\text{eval}}$ to be the $\lfloor p \cdot |D| \rfloor$ examples in $D$ with lowest value of $s(x)$, and choose $D_{\text{train}} = D \setminus D_{\text{eval}}$.
In \S\ref{sec:gpt-lm-setup}, we describe a few different ways to define $s$.
In \S\ref{sec:gpt-length}, we describe a modification to this procedure that controls for varying length between examples. 
Finally, we describe task-specific adjustments in \S\ref{sec:dset-choices}.

\subsection{Assigning Likelihood Scores $s(x)$}
\label{sec:gpt-lm-setup}

\begin{table*}[ht]
\centering
\footnotesize
\begin{tabular}{ l c c }
\toprule
\textbf{Task} & \textbf{Prompting} & \textbf{Fine-Tuning}\\
\midrule
\spider & \texttt{write a database question: \cbBlue{\{query\}}} & \texttt{<|endoftext|> \cbBlue{\{query\}}}\\
\midrule[\lightrulewidth]
\boolq & \multicolumn{2}{c}{\texttt{Passage: \{passage\} Ask a question about the passage: \cbBlue{\{question\}}}}\\
\midrule[\lightrulewidth]
\snli & \multicolumn{2}{c}{\texttt{Premise: \{premise\} This hypothesis is \{entailed/neutral/a contradiction\}: \cbBlue{\{hypothesis\}}}} \\
\bottomrule
\end{tabular}
\caption{\label{tab:task-input-formats}
Input formats for single-sentence and sentence-pair tasks in the prompting and fine-tuning settings. Values in curly braces are plugged in from the example. For \snli, we provide the label in the prompt to prime the LM to the class of hypothesis. The LM is trained (when fine-tuning) and evaluated on generating the query in \cbBlue{blue}.
}
\end{table*}

We use the total log-likelihood over the query tokens assigned by the \gpt language model as the score $s(x)$ for every example. There are two ways to use the LM: (1) prompting a frozen LM or (2) fine-tuning the LM on the dataset.

Past work has shown that prompting i.e. pre-pending a task-specific string to the query, helps \gpt generalize zero-shot to new tasks \citep{radford2019language-gpt2}. We use simple prompts that describe the task and prime the LM to the text we expect it to generate (see Table~\ref{tab:task-input-formats}). For sentence pair tasks (such as \snli and \boolq), it is necessary to compare the relation between two pieces of text and not just each piece in isolation. Thus, it is intuitive to describe unlikely examples by the conditional likelihood of $x_2$ given $x_1$. We demonstrate the flexibility of our approach by providing the label in the prompt if it adds additional information about the text to be generated (e.g. in \snli).\footnote{We include the label in the prompt for \snli but not \boolq because the resulting prompts seemed most natural for each dataset. This choice was made before assessing downstream behavior.}
We will refer to this setting which uses the prompted LM with the tag \emph{ll\_split pt} in the rest of the work.

The dataset curator may also choose to fine-tune the LM to better capture the task distribution.
We fine-tune the \gpt LM to maximize either the probability of $x$ for single sentence tasks or the conditional probability of $x_2$ given the prompt for sentence-pair tasks.
When fine-tuning the LM on the dataset, we need to ensure that it is not used to assign scores to the examples it is trained on. Given the dataset $D$, we first randomly partition $D$ into $k$ folds. For each fold, we fine-tune an LM on the remaining folds and use it to assign log-likelihood scores to examples in the held-out fold.
We refer the reader to Appendix~\ref{app:hyperparams} for fine-tuning details.
We will refer to this setting as \emph{ll\_split} henceforth.

\subsection{Controlling for Length}
\label{sec:gpt-length}
Since the likelihood of an utterance is negatively correlated with its length, we create a split that explicitly controls for the effect of length. After assigning a likelihood score to every utterance, the examples are bucketed based on length (defined  by tokenizing the utterance with NLTK \citep{10.3115/1118108.1118117-nltk}). For single-sentence and sentence-pair tasks, we use the length of the query ($x$ and $x_2$ respectively) over which log-likelihood was computed. Within each bucket, a fraction $p$ of the examples with the lowest $s(x)$ are put in the evaluation set; aggregating examples from all buckets, $| D_{\text{eval}} | \approx p \cdot |D|$.
We will refer to this control setting with the modifier \emph{(-len)} henceforth.\footnote{We also considered using perplexity, which normalizes for length, but it led to an over-correction where short examples were filtered into the evaluation set.}

\subsection{Dataset-specific Choices and Details}
\label{sec:dset-choices}

\paragraph{\spider.} We follow \citet{shaw-etal-2021-compositional-tmcd} and swap examples between the train and evaluation sets such that every logical program atom in the evaluation set appears at least once in the train set. This ensures that the model is not required to generalize to unseen function names and declarations.

\paragraph{\snli and \boolq.} We ensure label balance in our splits (as in the original data) by splitting the examples for each label separately, then combining the resulting train and evaluation sets.



\paragraph{Development sets.} \citet{csordas2021-devil-in-detail} show that without development sets that are in-distribution to challenging test sets, models are prone to over-fitting, which under-estimates their ability to generalize. Thus, after dividing the data into train and evaluation sets, we randomly divide the evaluation set into a development set and test set. Other details are reported in Appendix~\ref{app:dset-stats}.

\section{Experiments}
\label{sec:experiments}

Next, we benchmark task models on our Likelihood Splits. Splits created using \gptm will be the focus of our analysis. We will briefly study the effect of switching the LM to \gptl in \S\ref{sec:spider-gptl}. 

When creating Likelihood Splits, the number of folds $k$ for fine-tuning the LM (\S\ref{sec:gpt-lm-setup}) can be chosen by the dataset curator. For results in \S\ref{sec:experiments} and \S\ref{sec:split-properties}, we set $k=3$ arbitrarily. We analyse the effect of choosing a different value of $k$
in Appendix~\ref{app:ll-split-k-effect}. Our results show that the trends and observations discussed here hold true for other values of $k$.

\subsection{Benchmarked Models}

One of the goals of this work is to expose long-tail generalization as a challenge to state-of-the-art models; SotA models on the considered benchmarks are all pre-trained models. We make efforts to show that models with different pre-training data and objectives are similarly affected by our proposed splits. Hyperparameters and training details for the reported models are in Appendix~\ref{app:hyperparams}.

\paragraph{Semantic parsing.} Following \citet{shaw-etal-2021-compositional-tmcd}, we benchmark the competitive \tfive-base model \cite{raffel2020-t5} on all splits of the \spider dataset.
In order to test whether these splits are adversarial to the data splitting language model, we additionally fine-tune \gptm models for the semantic parsing task. To study the effect of model size, we fine-tune \tfive-small and \gpts variants.

\paragraph{\snli and \boolq.} We fine-tune two competitive models (\roberta \citep{liu2019roberta} and \electra \citep{clark2020electra}) at two model sizes (\textit{base} and \textit{large}).
Additionally, following \citet{poliak-etal-2018-hypothesis-spurious-nli}, we train a \roberta-large model to perform the task given just the hypothesis. The performance of a hypothesis-only model estimates the degree of spurious correlations that exist in the dataset which give away the label.

\subsection{Alternative Splits for Semantic Parsing}
\label{sec:spider-alternative-splits}

We compare the difficulty of the Likelihood Splits with past work on heuristic challenges splits.

\paragraph{Length.} Past work has established that text generation models trained on short inputs struggle to generalize to longer inputs at test time \citep{SCAN-dataset,PCFG-Set-Dataset,newman-etal-2020-eos}. We create \emph{Length} splits by placing examples with the longest input queries in the evaluation set and the remaining examples in the training set.

\paragraph{TMCD.} Systematicity is the ability to compositionally derive the meaning of an utterance from the known meaning of its parts. Past work studying systematicity in semantic parsing has defined ``atoms" as the smallest constituents of the grammar (e.g. variables and function names) and ``compounds" as complex structures formed by composing atoms (e.g. multi-argument functions and nested function calls) \citep{CFQ-Dataset}. Following \citet{shaw-etal-2021-compositional-tmcd}, we create TMCD (Target Maximum Compound Divergence) splits of \spider by maximizing the divergence between the distributions of compounds in the train and evaluation sets.

\paragraph{Template.} These splits test the ability of parsers to generate unseen program templates (canonicalized programs formed by anonymizing all variable names and standardizing syntax). We group examples in the \spider dataset based on templates defined by \citet{finegan-dollak-etal-2018-improving}. To create the evaluation set, we randomly pick groups of examples till the target set size is reached; the remaining groups form the training set.

\subsection{Model Performance on Likelihood Splits}
\label{sec:model-performance}


\begin{table}[bt]
\centering
\footnotesize
\begin{tabular}{@{}l c c c c@{}}
\toprule
\multirow{2}{*}{\textbf{Split}} & \textbf{\tfive} & \textbf{\tfive} & \textbf{\gpt} & \textbf{\gpt} \\
& \textbf{base} & \textbf{small} & \textbf{medium}{$\scriptstyle(\Delta)$} & \textbf{small} \\
\midrule
Random & 78.6 & 75.2 & 69.3 $\scriptstyle(9.3)$ & 64.7 \\
Length & 50.0 & 44.5 & 39.9 $\scriptstyle(10.1)$ & 34.0 \\
Template & 60.1 & 60.0 & 51.4 $\scriptstyle(8.7)$ & 45.1 \\
TMCD & 66.2 & 64.1 & 56.2 $\scriptstyle(10)$ & 51.4 \\
\midrule[\heavyrulewidth]
\multicolumn{5}{c}{\textbf{Split LM: \gptm}} \\
\midrule
ll\_split & 66.0 & 64.2 & 57.2 $\scriptstyle(8.8)$ & 51.8 \\
ll\_split (-len) & 71.3 & 67.3 & 59.9 $\scriptstyle(11.4)$ & 57.3 \\
ll\_split pt & 60.6 & 59.7 & 50.9 $\scriptstyle(9.7)$ & 45.9 \\
ll\_split pt (-len) & 73.5 & 68.4 & 64.5 $\scriptstyle(9)$ & 58.3 \\
\midrule[\heavyrulewidth]  
\multicolumn{5}{c}{\textbf{Split LM: \gptl}} \\
\midrule
ll\_split & 61.8 & 61.8 & 53.7 $\scriptstyle(8.1)$ & 48.3 \\
ll\_split (-len) & 69.7 & 66.2 & 59.1 $\scriptstyle(10.6)$ & 54.8 \\
ll\_split pt & 63.0 & 58.3 & 51.4 $\scriptstyle(11.6)$ & 45.7 \\
ll\_split pt (-len) & 72.0 & 70.1 & 63.4 $\scriptstyle(8.6)$ & 57.5 \\
\bottomrule
\end{tabular}
\caption{\label{tab:spider-parsing-acc}
\spider: Exact sequence prediction accuracy for Likelihood Splits created by \gptm and \gptl, and other challenge splits. Likelihood Splits are more challenging than random splits while not being adversarial to \gptm. $\Delta$ marks the performance drop from \tfive-base to \gptm.}
\end{table}

\begin{table*}[tb]
\centering
\footnotesize
\begin{tabular}{l c c c c c c c c c c@{}}
\toprule
& \multicolumn{5}{c}{\textbf{\snli}} & \multicolumn{5}{c}{\textbf{\boolq}}\\
\cmidrule(lr){2-6} \cmidrule(lr){7-11}
 & \multirow{2}{*}{Random} & \multicolumn{2}{c}{ll\_split} & \multicolumn{2}{c}{ll\_split pt} & \multirow{2}{*}{Random} & \multicolumn{2}{c}{ll\_split} & \multicolumn{2}{c}{ll\_split pt} \\
\cmidrule(lr){3-4} \cmidrule(lr){5-6} \cmidrule(lr){8-9} \cmidrule(lr){10-11}
\textbf{System} & & & (-len) & & (-len) & & & (-len) & & (-len) \\
\midrule[\heavyrulewidth]
\roberta-base & 89.6 \scriptsize{$\pm$0.4} & 79.3 & 77.1 & 82.6 & 81.7 & 74.9 \scriptsize{$\pm$0.4} & 71.6 & 71.2 & 72.4 & 71.9 \\
\roberta-large & 90.5 \scriptsize{$\pm$0.5}  & 82.4 & 79.2 & 85.0 & 84.3 & 84.4 \scriptsize{$\pm$0.6} & 79.3 & 78.9 & 82.3 & 80.6 \\
\electra-base & 90.5 \scriptsize{$\pm$0.2} & 80.1 & 78.4 & 82.9 & 82.8 & 78.8 \scriptsize{$\pm$1.1} & 74.1 & 74.3 & 75.2 & 73.6 \\
\electra-large & 91.0 \scriptsize{$\pm$1.3} & 82.6 & 81.6 & 85.9 & 84.9 & 85.5 \scriptsize{$\pm$0.6} & 82.6 & 82.1 & 83.7 & 81.9 \\
\midrule[\heavyrulewidth]
\roberta-large \scriptsize{(Hypothesis-only)} & 70.2 \scriptsize{$\pm$0.3} & 64.6 & 64.6 & 67.2 & 69.6 & - & - & - & - & - \\
\midrule
Human Accuracy & 88.7 \scriptsize{$\pm$0.8} & 83.6 & 84.4 & 85.2 & 86.4 & - & - & - & - & - \\
\bottomrule
\end{tabular}
\caption{\label{tab:snli-acc}
\snli and \boolq: Accuracy for various splits and model sizes. Likelihood Splits lead to decreased model performance. Controlling for length further increases the difficulty.}
\end{table*}

In Table~\ref{tab:spider-parsing-acc}, we report exact match accuracy\footnote{This metric accounts for the fact that SQL statements are invariant to certain shuffling and change in variable names.} on the data splits using the \textsc{Spider} evaluation suite. For \snli and \boolq, we report the accuracy of benchmarked models in Table~\ref{tab:snli-acc}. We create 3 random splits and report mean and standard deviation of accuracy of models trained on each split.

\paragraph{Likelihood Splits are more challenging than random splits.} 
On \spider, for example, \tfive-base accuracy on \emph{ll\_split} is 12.6 points lower than the random split accuracy. Likelihood Splits lead to drops in performance that are comparable to the alternative challenge splits. 
Only Likelihood Splits focus on challenges derived from input language variation; we analyze these challenges in \S\ref{sec:spider-split-properties}.

On \snli and \boolq, Likelihood Splits are also more challenging than random splits. For example, \electra-large accuracy decreases by 8.4 points on \snli and 2.9 points on \boolq.
On \snli, the performance of the hypothesis-only baselines on Likelihood Splits is lower than that on the random splits, 
which indicates that our splits are less easily solved by modeling spurious statistical cues.

\paragraph{Controlling for length preserves challenging nature of splits.}  Likelihood is negatively correlated with length, so Likelihood Split test data contains longer examples.
On \spider, generalizing to longer utterances is challenging, so controlling for length makes the Likelihood Splits less challenging. However, these splits are still much more challenging than random splits. For \tfive-base, \emph{ll\_split (-len)} is 7.3 points harder and \emph{ll\_split pt (-len)} is 5.1 points harder than the random split. 
By controlling for length, we identify examples that are more challenging for other reasons (discussed in \S\ref{sec:spider-split-properties}).
Fitting the dataset distribution with a fine-tuned LM reduces the correlation between length and likelihood on \spider.
Accordingly, \emph{ll\_split pt} poses a stronger length generalization challenge than \emph{ll\_split}, and thus is more challenging:
\tfive-base accuracy drops by 18 points on \emph{ll\_split pt} compared with the random split. 

Conversely, for \snli and \boolq, controlling for length makes the Likelihood Splits slightly harder compared to their uncontrolled versions (\electra-large accuracy drops by 1\% from \emph{ll\_split} to \emph{ll\_split (-len)} on \snli, and by 0.5\% on \boolq). 
This suggests length is not a reason that Likelihood Splits are harder for these datasets.
Relatedly, \emph{ll\_split pt} is easier than \emph{ll\_split} here.

\paragraph{Likelihood Splits do not unfairly penalize the scoring LM.} The difference in accuracy between \tfive-base and \gptm are comparable across all splits ($\Delta$ in Table~\ref{tab:spider-parsing-acc}). This shows that the Likelihood Splits do not unfairly penalize \gptm, the model used to create the Likelihood Splits. Thus, benchmarks based on Likelihood Splits will be fairer to model class of the LM used.

\paragraph{Human accuracy is less affected.} 
We estimate human accuracy on the evaluation sets using the $\sim$10\% of examples that were annotated with 5 labels in the original \snli dataset. Human accuracy is at most 5.1\% lower on our proposed splits than on the random splits. Model performance drops more severely than the smaller drop in human accuracy; models that were previously superhuman are now worse than the estimated human performance (except for \electra-large on \emph{ll\_split pt}).
In comparison, adversarial filtering \citep{LeBras2020AdversarialFilters} has a larger drop in human accuracy from 88\% on the standard split to 78\% on their most challenging split.
Thus, our method does not as heavily emphasize mislabeled or ambiguous examples.

\subsection{Effect of the LM on Likelihood Splits.}
\label{sec:spider-gptl}

We study the effect changing the language model by using a \gptl model to create the Likelihood Splits of \spider. The log-likelihood scores assigned to the examples by \gptm and \gptl are highly correlated; Pearson correlation coefficient (r) between log-likelihood scores from fine-tuned models is $0.96$ while it is $0.99$ for the pre-trained models. Accounting for swapping of examples in order to meet the atom constraint, the evaluation sets differ in 16\% of examples in the \emph{ll\_split} setting, and 10\% of the examples in \emph{ll\_split pt} setting. \emph{ll\_split} is more challenging when using \gptl; \tfive-base accuracy drops an additional 4.2\% compared to the \emph{ll\_split} with \gptm. The other splits are comparable with accuracies differing by 1-2\% across all models (see Table~\ref{tab:spider-parsing-acc}). Thus, we expect splits created with different LMs to demonstrate similar characteristics.

\subsection{Are Reverse Likelihood Splits Difficult?}

\begin{table}[tb]
\centering
\footnotesize
\begin{tabular}{l c c c c c@{}}
\toprule
 & \multirow{2}{*}{Random} & \multicolumn{2}{c}{ll\_split} & \multicolumn{2}{c}{ll\_split reverse} \\
\cmidrule(lr){3-4} \cmidrule(lr){5-6}
\textbf{System} & & & (-len) & & (-len)\\
\midrule[\heavyrulewidth]
\multicolumn{6}{c}{\textbf{\spider}} \\
\cmidrule(lr){1-6}
\textbf{\tfive-base} & 78.6 & 66.0 & 71.3 & 83.9 & 81.5 \\
\midrule[\heavyrulewidth]
\multicolumn{6}{c}{\textbf{\snli}} \\
\cmidrule(lr){1-6}
\textbf{E-base} & 90.5 \scriptsize{$\pm$0.2} & 80.1 & 78.4 & 96.6 & 97.4 \\
\textbf{E-large} & 91.0 \scriptsize{$\pm$1.3} & 82.6 & 81.6 & 96.7 & 97.4 \\
\midrule[\heavyrulewidth]
\multicolumn{6}{c}{\textbf{\boolq}} \\
\cmidrule(lr){1-6}
\textbf{E-base} & 78.8 \scriptsize{$\pm$1.1} & 74.1 & 74.3 & 77.8 & 78.1 \\
\textbf{E-large} & 85.5 \scriptsize{$\pm$0.6} & 82.6 & 82.1 & 85.8 & 86.8 \\
\bottomrule
\end{tabular}
\caption{\label{tab:all-acc-rev}
Accuracy on \spider, \snli and \boolq when training on the unlikely (tail) queries and evaluating on the likely (head) queries (\emph{ll\_split reverse}). The model accuracy on the reverse splits are comparable to or higher than accuracy on the random split. This supports the claim that generalizing to rare instances is a significant challenge. (E-base and E-large are \electra models)}
\end{table}


We wish to test whether the decrease in task accuracy is driven by rarity of the instances or whether any likelihood based distribution shift is challenging. We test this hypothesis by creating a setting that requires generalizing from the tail of the distribution to the head. Using the same likelihoods as before, we create reverse splits where the more likely (head) of the distribution is used as the evaluation set instead of the unlikely (tail). From Table~\ref{tab:all-acc-rev}, we see that the accuracy of \electra-large on \snli increases from 81.6\% on the Likelihood Split to 96.7\% on the reversed split. For comparison, this is more than the accuracy on random splits of \snli (91\%). We see similar trends on \boolq and \spider where the reverse splits are as easy as or easier than the corresponding random splits. We conclude that generalizing specifically to the tail is what makes our splits difficult.

\section{Analysis of Data Splits}
\label{sec:split-properties}


In order to highlight the challenges posed by our proposed splits, we analyze how the development sets (to ensure unseen test sets) differ from the training sets in each split. Our splits require models to simultaneously excel at many different skills believed to be important for language understanding.

\subsection{Properties of the Proposed \spider Splits}
\label{sec:spider-split-properties}

\paragraph{TMCD-related properties and length.} Following \citet{shaw-etal-2021-compositional-tmcd}, we report atom and compound divergences of the various splits in Table~\ref{tab:parsing-divergence} of Appendix~\ref{app:spider-div-expl}. Divergence measures how much the distribution of atoms/compounds differs between the train and evaluation set. Our approach leads to splits with higher than random atom divergence, which shows that our split poses the challenge of \textbf{generalizing to rare atoms}. Similarly, a greater than random compound divergence emerges from the resulting split. This means that the split also requires some amount of \textbf{compositional generalization}. From Figure~\ref{fig:spi-inp-len-var} in Appendix~\ref{app:length-variation}, we see that log-likelihood preferentially puts the longer queries in the test set and the corresponding length variation is closer to that of the length split than the other splits. Hence, it naturally requires some aspect of \textbf{length generalization}. As expected, by controlling for length of the utterances, we can remove the challenge of length generalization.

\begin{figure}[htb]
\centering
\includegraphics[width=0.48\textwidth]{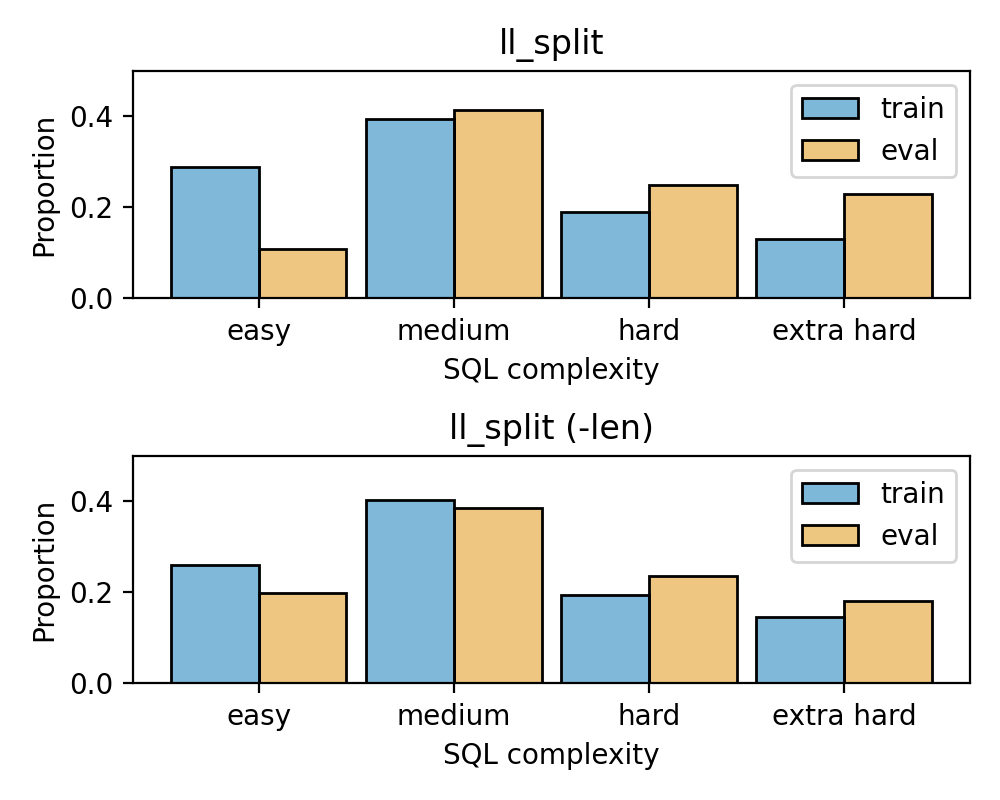}
\caption{\spider: Distribution of SQL programs of varying complexity in the train and development set of Likelihood Splits. These splits show a skew towards training on easy examples and evaluating on harder examples.}
\label{fig:spider-sql-hardness-main}
\end{figure}

\paragraph{Program difficulty.} \spider assigns a rating of `easy', `medium', `hard', or `extra hard' to every SQL program. From Figure~\ref{fig:spider-sql-hardness-main}, we see that the evaluation sets of the Likelihood Splits contain more examples from the harder categories than the training sets. 
Controlling for length reduces this effect but does not completely remove it (see Appendix~\ref{app:spider-sql-hardness} for more details). Note that this skew emerges even though we do not consider the programs when creating these splits.

\begin{figure}[htb]
\centering
\includegraphics[width=0.48\textwidth]{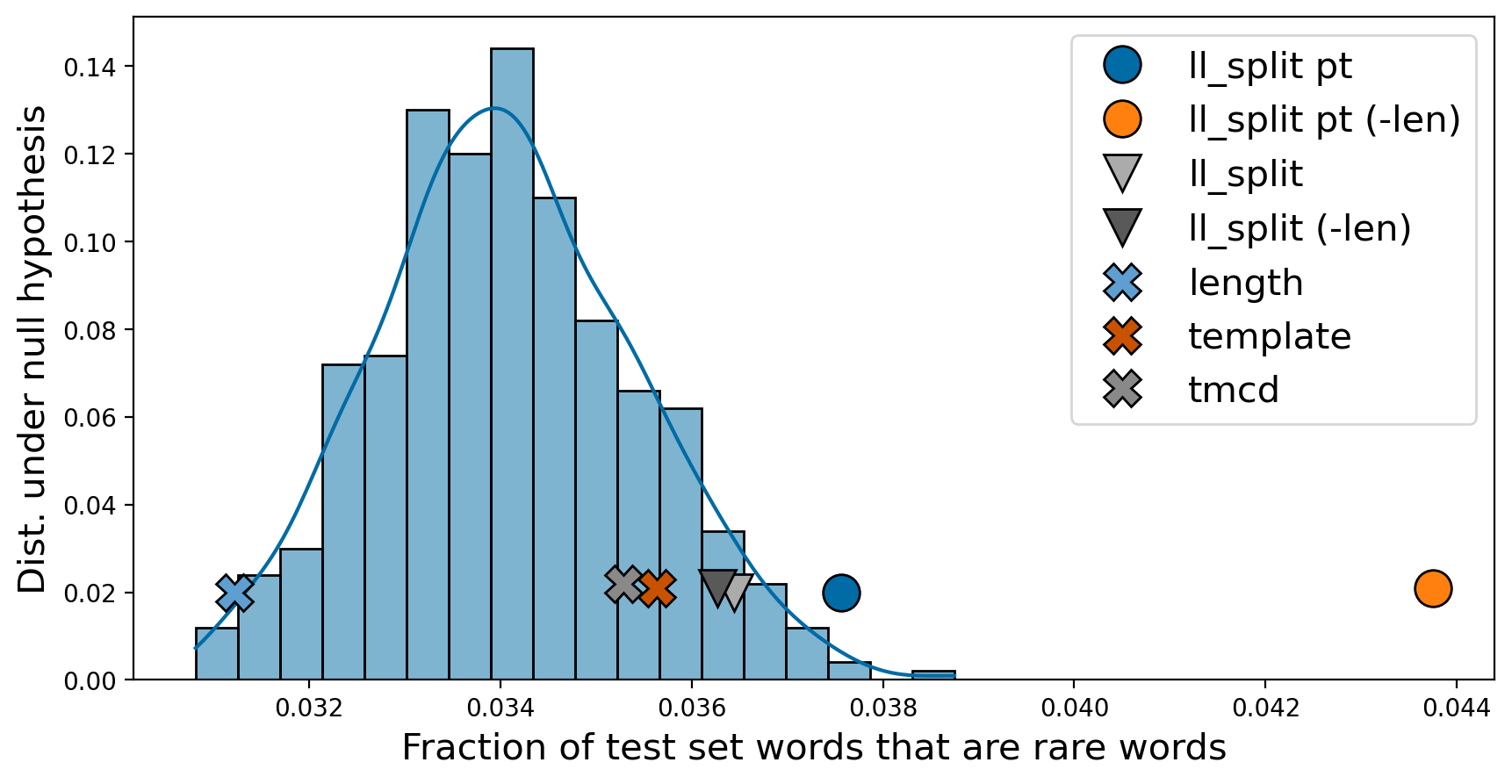}
\caption{\spider: Statistics of the fraction of dev set words that are rare for various splits. This is plotted against the distribution of values observed for 500 random slits of the data. ll\_split variants retain a larger fraction of rare words in the test set. Controlling for length finds shorter examples with more rare words.}
\label{fig:spider-rare-test-words}
\end{figure}

\paragraph{Rare words.} On the input side, we first analyze the distribution of rare words. We define rare words as all English words\footnote{We filter out incorrect spellings using the word list at \url{https://github.com/dwyl/english-words}} in the \spider dataset that occur at most 1 time per million words according to SUBTLEXus \cite{599801-subtlexus}. This results in a list of 561 words. We report the fraction of words in the development set that are rare. This metric automatically controls for the length of the examples; longer examples are more likely to contain rare words by chance.
To estimate the distribution of this fraction under random splits (null distribution), we create 500 random splits and plot the distribution of values observed. From Figure~\ref{fig:spider-rare-test-words}, we observe that the Likelihood Splits have more rare words in the test set, especially for the \emph{ll\_split pt} setting. Controlling for length puts shorter examples in the evaluation sets, but a larger fraction of the words are rare. The other challenging splits considered do not focus on the input language variation and hence the fraction of development set words that are rare is closer to random.

\paragraph{Input syntactic complexity.} We also study the query parse tree structures in various splits of the dataset in Figure~\ref{fig:spi-query-parse-stats}. We measure the complexity of the parse tree based on mean and max depth as well as Yngve score \citep{10.2307/985230-yngve} which is a measure of syntactic complexity. We see that more complex queries tend to be assigned lower likelihood and correspondingly put in the evaluation set. The effect of the complexity is also correlated with length and balancing for length reduces the gap between the complexity of the train and test set. We refer the reader to Appendix~\ref{app:spider-parse-variation} for more details.

\paragraph{Effect on accuracy.} In Appendix~\ref{app:spider-error-analysis} and Table \ref{tab:spi-err-sql-hard}, we show that the higher frequency of both novel compounds (i.e., compounds not seen during training) and harder programs each partially explain the higher difficulty of \emph{ll\_split pt} for \tfive-base.
For example, 16\% of dev examples in the random split have `extra hard' programs, compared with 25\% in \emph{ll\_split pt}. 
On the random split, \tfive-base gets 63\% of these examples correct, compared with 81\% dev accuracy overall, so these examples are indeed more challenging.
On `extra hard' examples in \emph{ll\_split pt}, \tfive-base has an even lower accuracy of 53\%. 
Thus, the mere fact that \emph{ll\_split pt} has more `extra hard' examples does not fully explain why it is harder; other factors must also be playing a role.

\begin{table}[t]
\centering
\footnotesize
\begin{tabular}{l c c}
\toprule
Category & Random & \emph{ll\_split pt} \\
\midrule
Easy        & 92.3\% (.225) & 81.3\% (.077) \\
Medium      & 82.3\% (.409) & 71.6\% (.435) \\
Hard        & 78.4\% (.201) & 60.1\% (.233) \\
Extra Hard  & 62.9\% (.164) & 52.5\% (.254) \\
\midrule
Dev set Acc & 80.6\% & 64.8\% \\
\midrule[\heavyrulewidth]
Projected Acc &  & 77.15\% \\
\bottomrule
\end{tabular}
\caption{\label{tab:spi-err-sql-hard} \spider: Accuracy of \tfive-base aggregated by the SQL hardness rating for random and \emph{ll\_split pt} dev sets. The number in brackets is the fraction of dev set examples that fall in each bucket. The examples in the dev set of \emph{ll\_split pt} are skewed towards harder examples. However, performance of \tfive-base on \emph{ll\_split pt} is lower than performance on random split in every bucket. Projecting and re-weighting the random set accuracies using the fraction of examples in each bucket in \emph{ll\_split pt} over-estimates dev set performance.
}
\end{table}


\subsection{Properties of the Proposed \snli Splits}
\label{sec:snli-split-properties}

For \snli, we study the variation of premise and hypothesis length (\ref{app:snli-length-variation}), distribution of rare words (\ref{app:snli-rare-test-words-variation}), Yngve score \citep{10.2307/985230-yngve} for syntactic complexity (\ref{app:snli-parse-variation}), and Flesch-Kincaid \citep{Kincaid1975DerivationON} reading grade-level (\ref{app:snli-reading-variation}). Evaluation sets of Likelihood Splits of \snli are more complex than their corresponding training sets on all 4 variations; evaluation set examples tend to be longer, tend to contain more rare words, are more syntactically complex, and have higher reading levels.

\begin{figure}[htb]
\centering
\includegraphics[width=0.35\textwidth]{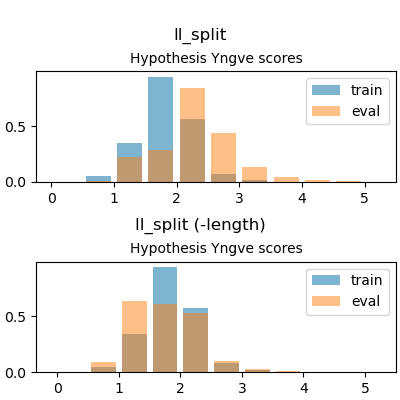}
\caption{\snli: Distribution of Yngve scores computed on the parse tree of the hypothesis. The evaluation sets for \emph{ll\_split} contain more complex utterances. Normalizing for the length surprisingly reverses the skew.}
\label{fig:snli-query-parse-stats-main}
\end{figure}

Controlling for length removes length variation, and slightly decreases the skew in reading level. Surprisingly, the Yngve scores of evaluation examples are skewed to being less complex than the corresponding training set (see Figure~\ref{fig:snli-query-parse-stats-main}), even though the length controlled variants of \snli are more challenging than the corresponding Likelihood Splits. Some the difficulty when controlling for length can be explained by the increased proportion of rare words.

We analyze the errors of \roberta-large on the development set of \emph{ll\_split (-len)}, the hardest \snli split (see Appendix~\ref{app:snli-err-analysis} for concrete examples). We find several instances of examples that require common-sense or world knowledge to be solved correctly. These include knowledge of terms such as crowd-surfing and lincoln logs (a type of toy), and facts like zip-lining is an exciting activity. We find that a small fraction of the errors are caused by ambiguous or incorrect labels. There are a several instances of spelling mistakes, a few of which change the meaning of the sentence.

\section{Conclusion}

With the saturation of static, single-metric leaderboards, there is growing consensus for the development of holistic evaluation benchmarks. This includes evaluation of systems on aspects of performance beyond just single error rate on in-distribution data; aspects such as 
performance on out-of-distribution data \citep{linzen-2020-accelerate}, and evaluating generalizability, robustness and fairness \citep{ethayarajh-jurafsky-2020-utility}. In this work, we describe an approach to benchmark long-tail generalization, a necessary skill for NLP systems that truly understand language. We demonstrate the challenge posed by our splits to state-of-the-art models on several tasks; standard evaluation overestimates model performance on long-tail utterances. Instead of releasing a random split as the only metric on official benchmarks, our simple method can be used, for a wide range of tasks, to expose additional challenges in the collected data at no annotation cost. Benchmarking long-tail generalization, in this manner, can test model behavior on a broad set of generalization challenges, which may be missed by evaluations that test specific skills in isolation.

\section*{Limitations}


Evaluating a proposed benchmarking method such as ours is challenging, as there is no community consensus on what properties characterize an ideal benchmark.
While we have argued that Likelihood Splits have a number of desirable properties, ultimately we intend Likelihood Splits to \emph{complement} other options for creating benchmarks, not replace them.
In particular, we do not aim to replace methods that require additional annotation and domain knowledge discussed in \S\ref{sec:related-work}. In situations where previously collected datasets contain no or very few examples of a particular type, creating new data may be the only way to test models on that type of example. We view our approach as one lightweight option that dataset curators can choose to create a more holistic benchmark.


The properties of the Likelihood Splits that we have studied in this work do not fully explain what makes the Likelihood Splits harder. Dataset splits that explicitly test specific skills like length generalization and compositional generalization are good at exposing specific weaknesses in models. While it is hard to pinpoint the source of difficulty, our approach is complementary in that it can test a much broader set of skills that a narrow test may miss.

The difficulty of out-of-distribution generalization is higher in low resource languages, however, we show that the problem is yet not solved for NLP tasks even in the high resource English language. Our approach has the flexibility to use any autoregressive LM to score the utterances; large multi-lingual LMs such as BLOOM \citep{https://doi.org/10.48550/arxiv.2211.05100-bloom} can be used if appropriate.

The model performance gaps between random split and Likelihood Splits are small (2-4\%) on some datasets (e.g. \boolq). We cannot guarantee that Likelihood Splits for a new dataset will be much more challenging than random splits. In such a situation, other complementary evaluation strategies may be recommended to more strenuously challenge models.


Our approach has multiple knobs to control the properties of the splits created: (1) prompting/fine-tuning the LM, (2) controlling length variation, and (3) dataset specific choices such as label balancing. This choice gives dataset curators a lot of control to modify the approach. It is possible that the behaviour of the splits might be inconsistent under some changes. In our experiments, we find that qualitative findings are largely consistent, even across changes such as using a different language model.

Finally, there is no guarantee that the challenge posed is a fair generalization task \citep{geiger-etal-2019-posing-fair-generalization-tasks}; we cannot guarantee that all skills needed to solve the test set can be learned from the training set. Nevertheless, since our approach partitions data that was collected under a single consistent protocol, it is more likely to be fair than methods that rely on an additional, separate annotation process to create test data.





\section*{Acknowledgements}

We thank Xiang Ren for discussions and feedback in the initial stages of the work. We thank Rajarshi Das and colleagues at USC for feedback on drafts of this work. This work was funded in part by a gift from Open Philanthropy.

\bibliography{custom}
\bibliographystyle{acl_natbib}

\appendix
\section{Appendices}

\subsection{Dataset Statistics}
\label{app:dset-stats}

Refer to Table~\ref{tab:dset-stats} for final split sizes. When creating the splits, we first partition the available data into train and evaluation (combined size of dev + test) sets using the methodology of each split (e.g. TMCD maximizes compound divergence, Likelihood Splits sort by an LM score and then partition the data). Then the evaluation set is randomly divided into dev and test sets.

\begin{table}[htb]
\centering
\footnotesize
\begin{tabular}{l c c c c}
\toprule
Dataset & |Train| & |Dev| & |Test| & Total\\
\midrule
\spider & 5,966      & 1,034  & 1,034  & 8,034\\
\snli   & 549,018    & 10,000 & 10,000 & 569,018 \\
\boolq  & 7,617      & 2,540  & 2,540  & 12,697 \\
\bottomrule
\end{tabular}
\caption{\label{tab:dset-stats} Sizes of train/dev/test sets for the dataset splits
}
\end{table}

The public release of the \spider dataset consists of 7000 training examples and 1034 validation examples (it also contains 1659 additional examples from older datasets which we do not use in our work). We use these 8034 examples to create all our splits. \citet{shaw-etal-2021-compositional-tmcd}, one of the alternative splits that we compare against, use a subset of 4000 examples from the 7000 training examples. Hence, our results are not directly comparable to performance reported by them. The SQL programs for 6 (of 8034) examples in the dataset cannot be parsed uniquely and thus we cannot define compounds on these examples. We drop these examples when creating the TMCD split i.e. the training set of TMCD split contains 6 fewer examples.

The \snli data contains 550152 training examples, 10000 dev examples and 10000 test examples for a total of 570152 examples. We drop examples where the gold label cannot be determined by majority vote. We also drop examples where the premise was labelled `Cannot see picture to describe.' or the hypothesis is empty. This results in a filtered dataset of 569018 examples from which we create our splits.

The public release of \boolq contain 9427 labeled training examples, 3270 labeled development examples, and 3245 unlabeled test examples. Thus we create our splits from the 12,697 labelled train and development examples. We maintain the approximate 60/20/20 train/dev/test proportions of the original dataset when creating the splits.

\subsection{Model Hyperparameters}
\label{app:hyperparams}

We use the Transformers library \cite{wolf-etal-2020-transformers-huggingface} for training and evaluation. All models were trained on Nvidia Quadro RTX 6000 GPUs (24GB GPU Memory).

We report hyperparameters for fine-tuning \gpt to create the Likelihood Splits in Table~\ref{tab:gpt-lm-hyper}. We select the best checkpoint based on lowest perplexity by validating on 10\% of the training data in each fold.

\begin{table}[t]
\centering
\footnotesize
\begin{tabular}{@{}l c@{}}
\toprule
Hyperparameter & Value \\
\midrule
train batch\_sz & 32 \\
lr\_scheduler & constant \\
learning\_rate & 2e-5 \\
optimizer & AdamW \\
eval steps & 64 \\
\midrule[\lightrulewidth]
\multirow{3}{*}{max steps} & \spider: 2000 \\
 & \snli: 15000 \\
 & \boolq: 3000 \\
\bottomrule
\end{tabular}
\caption{\label{tab:gpt-lm-hyper} Hyperparameters for fine-tuning \gpt (both medium and large) on the dataset folds to create Likelihood Splits
}
\end{table}

Hyperparameters for \spider models are in Table~\ref{tab:spider-hyper}, for \snli models in Table~\ref{tab:snli-hyper} and for \boolq models in Table~\ref{tab:boolq-hyper}. Note that we evaluate checkpoints (see hyperparameter `eval steps') during training to select the best checkpoint at the end of training.

For \spider, we follow \citet{shaw-etal-2021-compositional-tmcd} and tune the learning rate, batch size and maximum training steps for a \tfive-base model \cite{raffel2020-t5} on a random split of the \spider dataset. Once we have found a hyperparameter setting, we apply the same setting on the all splits. We also report performance of a \tfive-small model on all splits trained with the same hyperparameters.

For \snli and \boolq, we follow the default hyperparameters suggested by the original works. Additionally, we perform early stopping when performance on the validation set fails to improve for a specified number of evaluations.

\begin{table}[t]
\centering
\footnotesize
\begin{tabular}{@{}l  c  c@{}}
\toprule
 & \small{\textbf{\tfive}} & \small{\textbf{\gpt}} \\
\midrule
train batch\_sz & 8 & 2 \\
grad acc steps & 16 & 16 \\
max\_steps & 10000 & 10000 \\
lr\_scheduler & constant & constant \\
learning\_rate & 1e-3 & 2e-5 \\
optimizer & Adafactor & AdamW \\
max src\_length & 512 & 512\\
max tgt\_length & 256 & 256 \\
\midrule[\lightrulewidth]
\multirow{3}{*}{src prefix} & \multirow{3}{*}{-} & \texttt{database} \\
 & & \texttt{question for} \\
 & & \texttt{table } \\
\midrule[\lightrulewidth]
\multirow{2}{*}{tgt prefix} & \multirow{2}{*}{\texttt{semanticparse: }} & \texttt{generate} \\
 & & \texttt{the sql parse: } \\
\midrule[\lightrulewidth]
eval batch\_sz & 8 & 1 \\
eval steps & 256 & 128 \\
num\_beams & 5 & 5 \\
\bottomrule
\end{tabular}
\caption{\label{tab:spider-hyper} Hyperparameters for the models trained on \spider.
}
\end{table}

\begin{table}[th]
\centering
\footnotesize
\begin{tabular}{l c c c}
\toprule
 & \multirow{2}{*}{\small{\textbf{\roberta}}} & \multicolumn{2}{c}{\small{\textbf{\electra}}} \\
\cmidrule{3-4}
 & & \small{\textbf{base}} & \small{\textbf{large}} \\
\midrule
train batch\_sz     & \multicolumn{3}{c}{32} \\
max seq length      & \multicolumn{3}{c}{128} \\
lr\_scheduler       & \multicolumn{3}{c}{linear} \\
optimizer           & \multicolumn{3}{c}{AdamW} \\
adam\_beta1         & \multicolumn{3}{c}{0.9} \\
adam\_beta2         & 0.98 & \multicolumn{2}{c}{0.999}\\
adam\_epsilon       & \multicolumn{3}{c}{1e-6} \\
num epochs          & 10 & \multicolumn{2}{c}{3}\\
warmup ratio        & 0.06 & \multicolumn{2}{c}{0.1} \\
layer. lr decay     & 1.0 & 0.8 & 0.9 \\
learning\_rate      & 1e-5 & 1e-4 & 5e-5 \\
weight decay        & 0.1 & \multicolumn{2}{c}{0.0}\\
\midrule[\lightrulewidth]
eval batch\_sz      & \multicolumn{3}{c}{32}\\
eval steps          & \multicolumn{3}{c}{256} \\
patience            & 20 & \multicolumn{2}{c}{n/a} \\
\bottomrule
\end{tabular}
\caption{\label{tab:snli-hyper} Hyperparameters for the models trained on \snli. Patiences refer to number of evaluations with no improvement before early stopping.
}
\end{table}

\begin{table}[th]
\centering
\footnotesize
\begin{tabular}{l c c c}
\toprule
 & \multirow{2}{*}{\small{\textbf{\roberta}}} & \multicolumn{2}{c}{\small{\textbf{\electra}}} \\
\cmidrule{3-4}
 & & \small{\textbf{base}} & \small{\textbf{large}} \\
\midrule
train batch\_sz     & \multicolumn{3}{c}{8} \\
grad acc steps      & \multicolumn{3}{c}{4} \\
max seq length      & \multicolumn{3}{c}{512} \\
lr\_scheduler       & \multicolumn{3}{c}{linear} \\
optimizer           & \multicolumn{3}{c}{AdamW} \\
adam\_beta1         & \multicolumn{3}{c}{0.9} \\
adam\_beta2         & 0.98 & \multicolumn{2}{c}{0.999}\\
adam\_epsilon       & \multicolumn{3}{c}{1e-6} \\
num epochs          & 10 & \multicolumn{2}{c}{5}\\
warmup ratio        & 0.06 & \multicolumn{2}{c}{0.1} \\
layer. lr decay     & 1.0 & 0.8 & 0.9 \\
learning\_rate      & 1e-5 & 1e-4 & 5e-5 \\
weight decay        & 0.1 & \multicolumn{2}{c}{0.0}\\
\midrule[\lightrulewidth]
eval batch\_sz      & \multicolumn{3}{c}{8}\\
eval steps          & \multicolumn{3}{c}{128} \\
patience            & 10 & \multicolumn{2}{c}{n/a} \\
\bottomrule
\end{tabular}
\caption{\label{tab:boolq-hyper} Hyperparameters for the models trained on \boolq. Patience refers to number of evaluations with no improvement before early stopping.
}
\end{table}

\subsection{Effect of $k$ on Likelihood Splits}
\label{app:ll-split-k-effect}

\begin{table}[htb]
\centering
\footnotesize
\begin{tabular}{l c c c c c@{}}
\toprule
& \multicolumn{5}{c}{\textbf{\spider}} \\
\cmidrule(lr){2-6}
 & \multirow{2}{*}{Random} & \multicolumn{2}{c}{ll\_split (k=3)} & \multicolumn{2}{c}{ll\_split (k=5)} \\
\cmidrule(lr){3-4} \cmidrule(lr){5-6}
\textbf{System} & & & (-len) & & (-len)\\
\midrule[\heavyrulewidth]
\textbf{\tfive-base} & 78.6 & 66.0 & 71.3 & 64.8 & 69.2 \\
\bottomrule
\end{tabular}
\caption{\label{tab:spider-k-effect}
Effect of $k$ on the difficulty of Likelihood Splits of \spider. The accuracies of \tfive-base on the Likelihood Split are comparable and significantly lower than the accuracy on Random split. Controlling for length decreases the difficulty in both cases.}
\end{table}

\begin{table*}[tb]
\centering
\footnotesize
\begin{tabular}{l c c c c c c c c c c@{}}
\toprule
& \multicolumn{5}{c}{\textbf{\snli}} & \multicolumn{5}{c}{\textbf{\boolq}}\\
\cmidrule(lr){2-6} \cmidrule(lr){7-11}
 & \multirow{2}{*}{Random} & \multicolumn{2}{c}{ll\_split (k=3)} & \multicolumn{2}{c}{ll\_split (k=5)} & \multirow{2}{*}{Random} & \multicolumn{2}{c}{ll\_split (k=3)} & \multicolumn{2}{c}{ll\_split (k=5)} \\
\cmidrule(lr){3-4} \cmidrule(lr){5-6} \cmidrule(lr){8-9} \cmidrule(lr){10-11}
\textbf{System} & & & (-len) & & (-len) & & & (-len) & & (-len) \\
\midrule[\heavyrulewidth]
\roberta-base & 89.6 \scriptsize{$\pm$0.4} & 79.3 & 77.1 & 78.4 & 75.0 & 74.9 \scriptsize{$\pm$0.4} & 71.6 & 71.2 & 71.1 & 70.3 \\
\roberta-large & 90.5 \scriptsize{$\pm$0.5} & 82.3 & 79.2 & 82.1 & 79.0 & 84.4 \scriptsize{$\pm$0.6} & 79.3 & 78.9 & 78.4 & 78.3 \\
\electra-base & 90.5 \scriptsize{$\pm$0.2} & 80.1 & 78.3 & 80.1 & 77.6 & 78.8 \scriptsize{$\pm$1.1} & 74.1 & 74.3 & 75.0 & 73.9 \\
\electra-large & 91.0 \scriptsize{$\pm$1.3} & 82.6 & 81.6 & 84.0 & 80.6 & 85.5 \scriptsize{$\pm$0.6} & 82.6 & 82.1 & 83.0 & 80.6 \\
\midrule
Human Accuracy & 88.7 \scriptsize{$\pm$0.8} & 83.6 & 84.4 & 83.0 & 84.0 & - & - & - & - & - \\
\bottomrule
\end{tabular}
\caption{\label{tab:snli-boolq-k-effect}
Effect of $k$ on the difficulty of Likelihood Splits of \snli and \boolq. There are some accuracy differences on \boolq, however the values are comparable and significantly lower than the accuracy on Random splits. The accuracy differences are less pronounced on \snli.}
\end{table*}

When creating Likelihood Splits, the number of folds $k$ for fine-tuning (\S\ref{sec:gpt-lm-setup}) is a choice left to the creator of the benchmark. In our work, we set $k=3$ as an arbitrary choice prior to running the task models i.e. it was not tuned based on task model performance. We conduct additional experiments to test the effect of changing the value of $k$ by setting $k=5$ and generating new splits.

On \spider, when $k=5$ instead of $k=3$, Likelihood scores are highly correlated with a Pearson's r of 0.90. However, the process of balancing atoms (described in \S\ref{sec:dset-choices}) causes the evaluation sets to look more different. When controlling for length, 77\% of the evaluation set examples are the same; 80\% of the evaluation examples are the same otherwise. We report the accuracy of \tfive-base (the most competitive baseline model on \spider) on the new splits with $k=5$ in Table~\ref{tab:spider-k-effect}. The new splits are more difficult by about 2\%. We observe the same trend where controlling for length makes the splits less challenging.

On \boolq, when using 5 folds instead of the 3 folds, Likelihood scores are highly correlated with a Pearson's r of 0.96 and 89\% of the evaluation set examples are the same. Accordingly, the ELECTRA-large accuracy only changes slightly from 82.6\% to 83\% on the new test set and is still lower than the random split accuracy. While there exists an indication that controlling for length makes the \emph{ll\_split (-len)} splits more challenging, the effect of controlling for length becomes more pronounced when $k=5$. We report the effect of $k$ on \boolq performance in detail in Table~\ref{tab:snli-boolq-k-effect}. 

We report the effect of changing $k$ on \snli accuracy in Table~\ref{tab:snli-boolq-k-effect}. Since the \snli dataset is an order of magnitude larger than the \spider and \boolq datasets, the number of folds has less of an impact on the LM fine-tuning. As a result, Likelihood scores for $k=3$ and $k=5$ are highly correlated with a Pearson's r of 0.99. When controlling for length, 89\% of the evaluation set examples are the same; 92\% of the evaluation examples are the same otherwise. Accordingly, we see much smaller differences in model performance on the new splits; the \roberta model accuracies change by at most 0.9\% on the new test sets.

\subsection{\spider: Variation of TMCD Related Properties}
\label{app:spider-div-expl}

Past work by \citet{CFQ-Dataset} has established the terms of atom and compound ``divergence" to quantitatively describe the extent to which the distributions of the atoms and compounds differ between the train and evaluation sets. They used the Chernoff coefficient \citep{CHUNG1989280-chernoff-coefficient} to measure distribution similarity:

\begin{equation}
C_\alpha ( P \parallel Q ) = \sum_k p_k^{\alpha} q_k^{1 - \alpha} \qquad\in [0, 1]
\end{equation}

where $p_k$ and $q_k$ are the probability of a particular atom/compound $k$ being in the train and test set respectively. The divergence is then $1 - C_\alpha$. The ``atom" divergence uses $\alpha = 0.5$ as a symmetric divergence score. The ``compound" divergence used $\alpha = 0.1$ to give more importance to the occurrence of a compound in the train set rather than trying to make the distributions of train and test set similar.

\begin{table}[ht]
\centering
\small
\begin{tabular}{ l c c }
\toprule
\textbf{Split} & \textbf{Atom} & \textbf{Compound}\\
\midrule
Random & 0.077 & 0.046 \\
Length & 0.120 & 0.092 \\
Template & 0.105 & 0.089 \\
TMCD & 0.296 & 0.322 \\
\midrule
ll\_split & 0.083 & 0.054 \\
ll\_split (-len) & 0.081 & 0.049 \\
\midrule
ll\_split prompt & 0.093 & 0.056 \\
ll\_split prompt (-len) & 0.094 & 0.052 \\
\bottomrule
\end{tabular}
\caption{\label{tab:parsing-divergence}
Atom and Compound divergence (on the logical form side) between train and dev sets of various splits. Although we ensure every atom appears at least once in the train set, a high atom divergence demonstrates the challenge of learning rare atoms. A greater than random compound divergence emerges denoting a need for compositional generalization.
}
\end{table}

\subsection{\spider: Variation of SQL Hardness}
\label{app:spider-sql-hardness}

\begin{figure}[htb]
\centering
\includegraphics[width=0.47\textwidth]{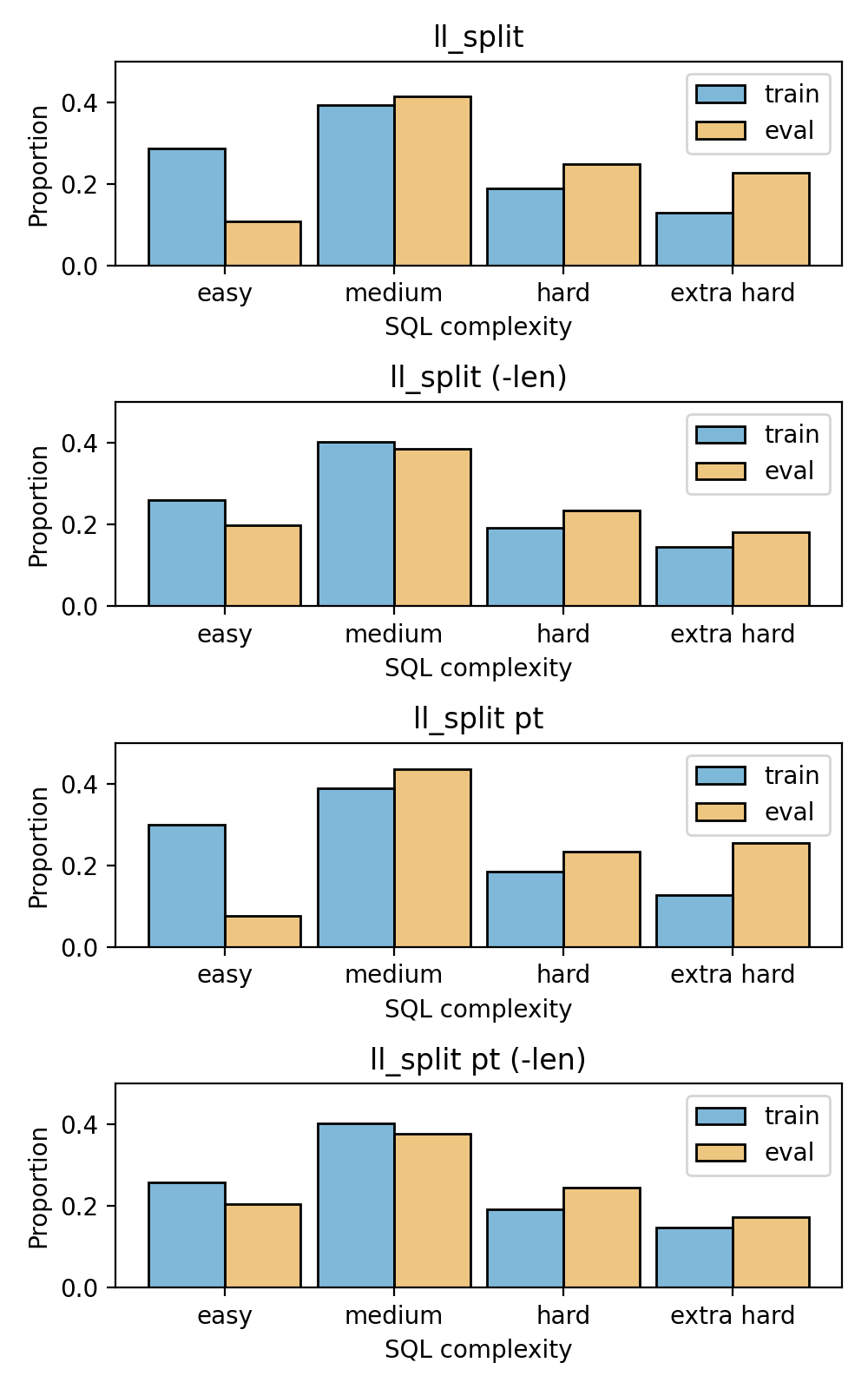}
\caption{\spider: Distribution of SQL programs of varying complexity in the train and development set of various splits. Likelihood Splits show a skew towards training on easy examples and evaluating on harder examples.}
\label{fig:spider-sql-hardness-var}
\end{figure}

We use a tool provided by the \spider dataset creators to evaluate hardness. The tools assigns a rating from easy, medium, hard or extra hard to every example based on the complexity of the SQL parse. Complexity is evaluated in terms of the number of join and aggregation operations, and nested SQL statements. We find that the Likelihood Splits are skewed towards putting more complex examples in the evaluation set compared to the test set (see Figure~\ref{fig:spider-sql-hardness-var}).

\subsection{\spider: Input Length Variation}
\label{app:length-variation}

\begin{figure}[htb]
\centering
\includegraphics[width=0.47\textwidth]{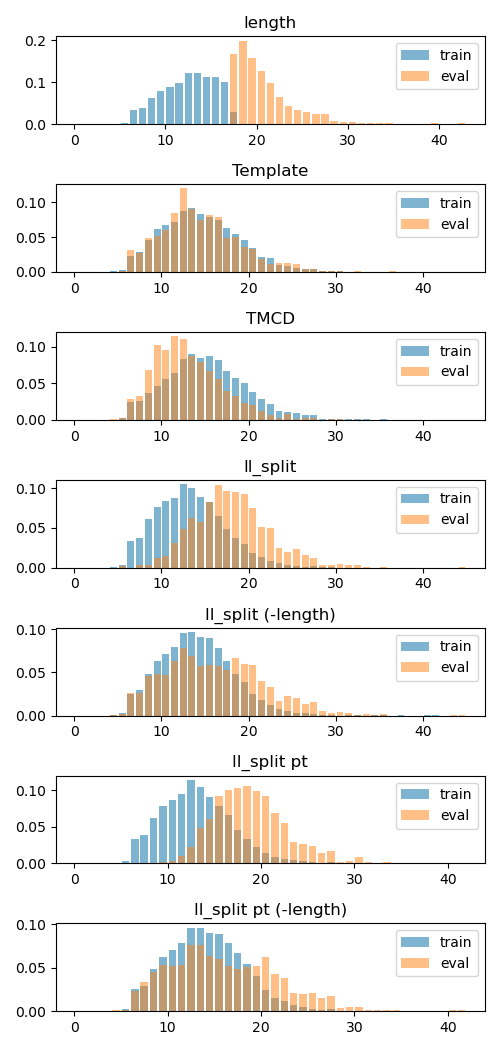}
\caption{\spider: Input length variation for the splits. Y-axis is the distribution of examples within each length bucket of the X-axis}
\label{fig:spi-inp-len-var}
\end{figure}

As expected, the likelihood assigned by the language model (LM) is negatively correlated with sequence length meaning i.e. longer sequences tend to have lower likelihood. This can be seen from Figure~\ref{fig:spi-inp-len-var}, where \emph{ll\_split} and \emph{ll\_split pt} tend to put longer utterances in the lower likelihood evaluations set. Accounting for length by dividing the data within buckets makes the distribution of train and test sets align better and remove the added difficulty of length generalization. The \emph{length} split poses this challenge which has been established to be a difficult ability for generation models to acquire \citep{newman-etal-2020-eos}. Note that the distributions do not match exactly since examples need to be swapped between train and evaluation set to meet the atom constraint (evaluation cannot contain any unseen atoms).

\begin{figure}[htb]
\centering
\includegraphics[width=0.45\textwidth]{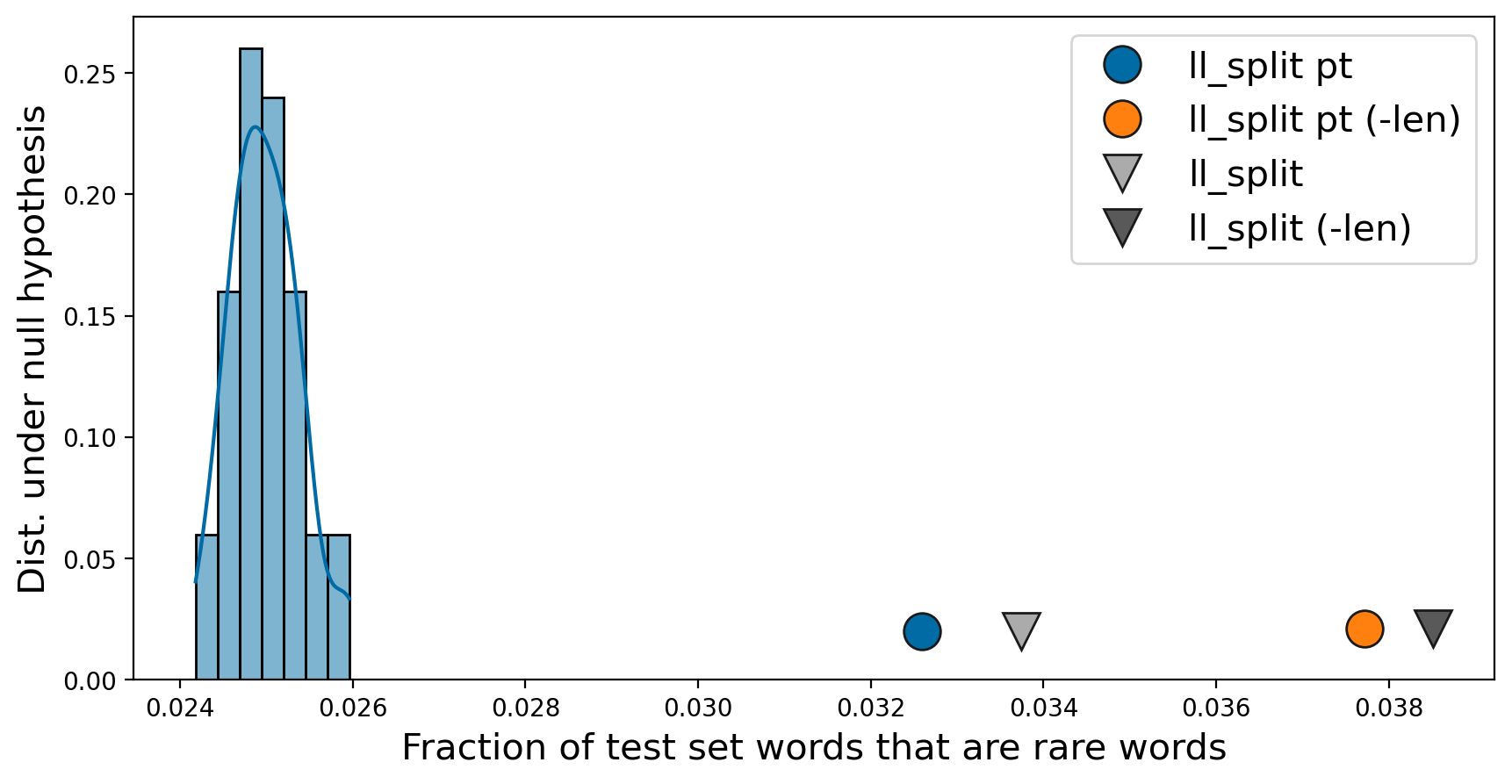}
\caption{\snli: Fraction of development set words that are rare in the premise and hypothesis of various splits. The dev sets for \emph{ll\_split} seem to contain a larger fraction of rare words than random splits. Normalizing for the length seems to retain more rare words.}
\label{fig:snli-rare-test-words}
\end{figure}

\subsection{\spider: Variation of Query Parse Structure}
\label{app:spider-parse-variation}

We analyze the complexity of the parse structure of the queries. Following \citet{https://doi.org/10.48550/arxiv.2110.08514-analyzing-dadc}, we parse the queries using the Benepar parser \citep{kitaev-klein-2018-constituency-benepar} based on T5 small \citep{raffel2020-t5}. We report the distributions of mean and max parse tree depth as well as the syntactic complexity of the utterance based on the Yngve score \citep{10.2307/985230-yngve,roark-etal-2007-syntactic}. The Yngve score essentially measures the average number of left branches on the path from the root of the parse tree to every word in the sentence and can be thought of as measuring the number of spans that need to be coordinated.

\begin{figure*}[htb]
\centering
\includegraphics[width=0.9\textwidth]{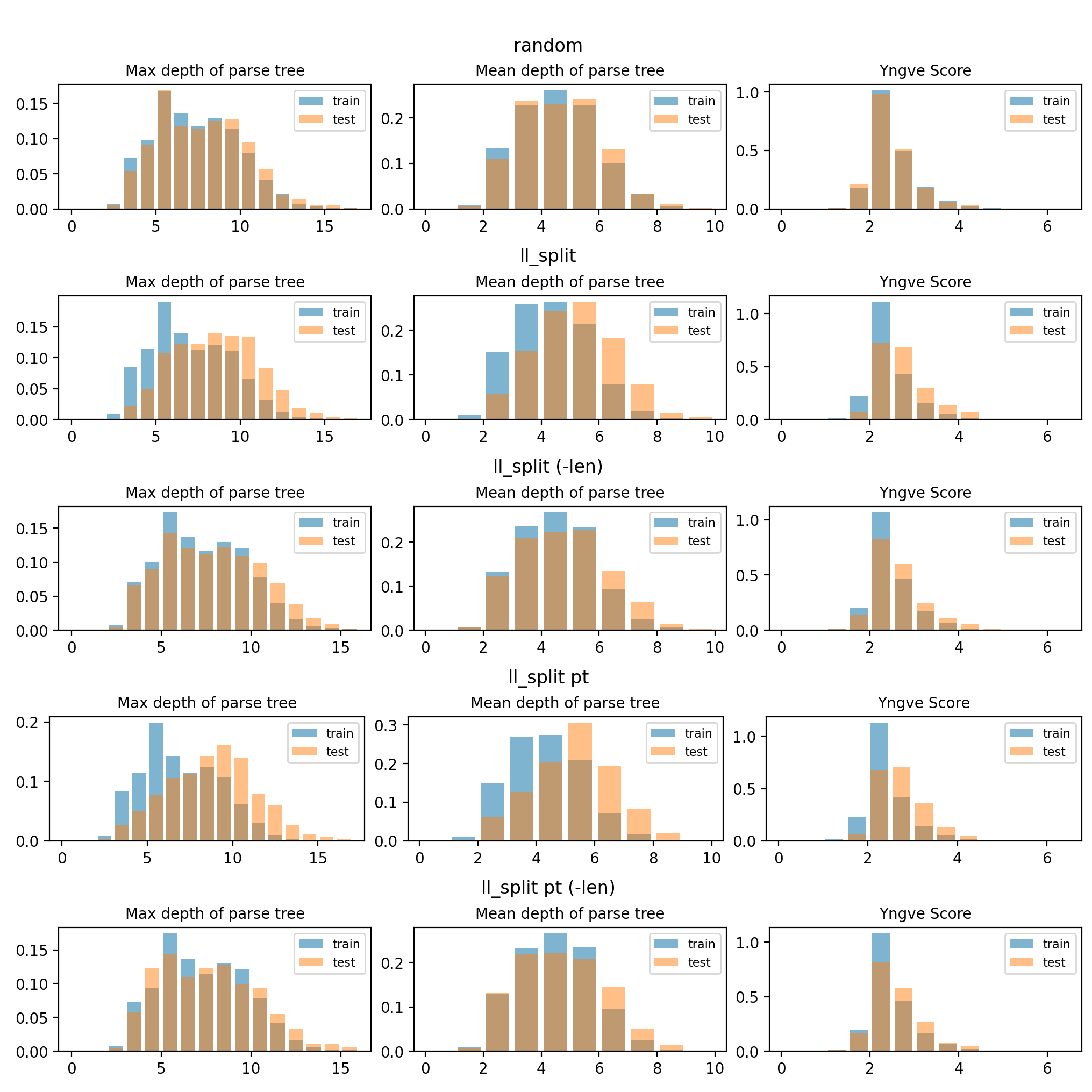}
\caption{\spider: Distribution of features computed on the parse tree of the input query. The dev sets for \emph{ll\_split} seem to contain more complex utterances across all 3 metrics considered. Normalizing for the length seems to reduce the effect. The metrics are mean and max depth of parse tree and Yngve score which is a metric}
\label{fig:spi-query-parse-stats}
\end{figure*}

We can see that the dev set of the \emph{ll\_split} is on average more complex than its train set along all 3 metrics considered (see Figure~\ref{fig:spi-query-parse-stats}). Moreover, these metrics are correlated with utterance length, and controlling for it in the \emph{ll\_split (-len)} split makes the difference less pronounced.

\subsection{\spider: Error Analysis}
\label{app:spider-error-analysis}

We analyze the performance of \tfive-base on the development set of \emph{ll\_split pt}. In particular, we test whether the presence of novel compounds and SQL query hardness are sufficient to explain the difficulty.

We call compounds in the SQL programs of the development set as `novel' if they do not occur in the training set of the split. 25.5\% of the dev set examples in the random split contain at least one novel compound as opposed to 43.6\% of the dev set examples of \emph{ll\_split pt}. From Table~\ref{tab:spi-err-novel-comp}, we see that \tfive-base performance is lower in both categories of examples. Projecting for expected performance on dev set of \emph{ll\_split pt} assuming the examples were as difficult as examples from the random split over-estimates the performance of \tfive-base.

\begin{table}[tb]
\centering
\footnotesize
\begin{tabular}{l c c}
\toprule
Split & Random & \emph{ll\_split pt} \\
\midrule
Percent. of examples & \multirow{2}{*}{25.5\%} & \multirow{2}{*}{43.6\%} \\
\qquad with a novel compound \\
Acc on examples & \multirow{2}{*}{61.4\%} & \multirow{2}{*}{45.5\%} \\
\qquad with novel compounds\\
Acc on remaining examples & 87.1\% & 79.8\% \\
\midrule
Acc on the dev set & 80.6\% & 64.8\% \\
\midrule[\heavyrulewidth]
Projected accuracy &  & 75.9\% \\
\bottomrule
\end{tabular}
\caption{\label{tab:spi-err-novel-comp} \spider: The presence of novel compounds alone does not explain the difficulty of the \emph{ll\_split pt}. Projecting the random set accuracies using the percentage of examples with novel compounds in \emph{ll\_split pt} over-estimates dev set performance.}
\end{table}

We report performance of \tfive-base on the dev sets grouped by the SQL hardness metric (described in Appendix~\ref{app:spider-sql-hardness}) in Table~\ref{tab:spi-err-sql-hard}. We see that accuracy on \emph{ll\_split pt} is lower than the accuracy on the random set within each SQL complexity bucket. If the sole source of difficulty was the larger proportion of harder examples, projecting the random set accuracies would correctly estimate dev set performance on \emph{ll\_split pt}. However, the projection is an over-estimate. Thus, the hardness metric alone does not explain the difficulty of the proposed split.

\subsection{\snli: Input Length Variation}
\label{app:snli-length-variation}

\begin{figure*}[htb]
\centering
\includegraphics[width=0.8\textwidth]{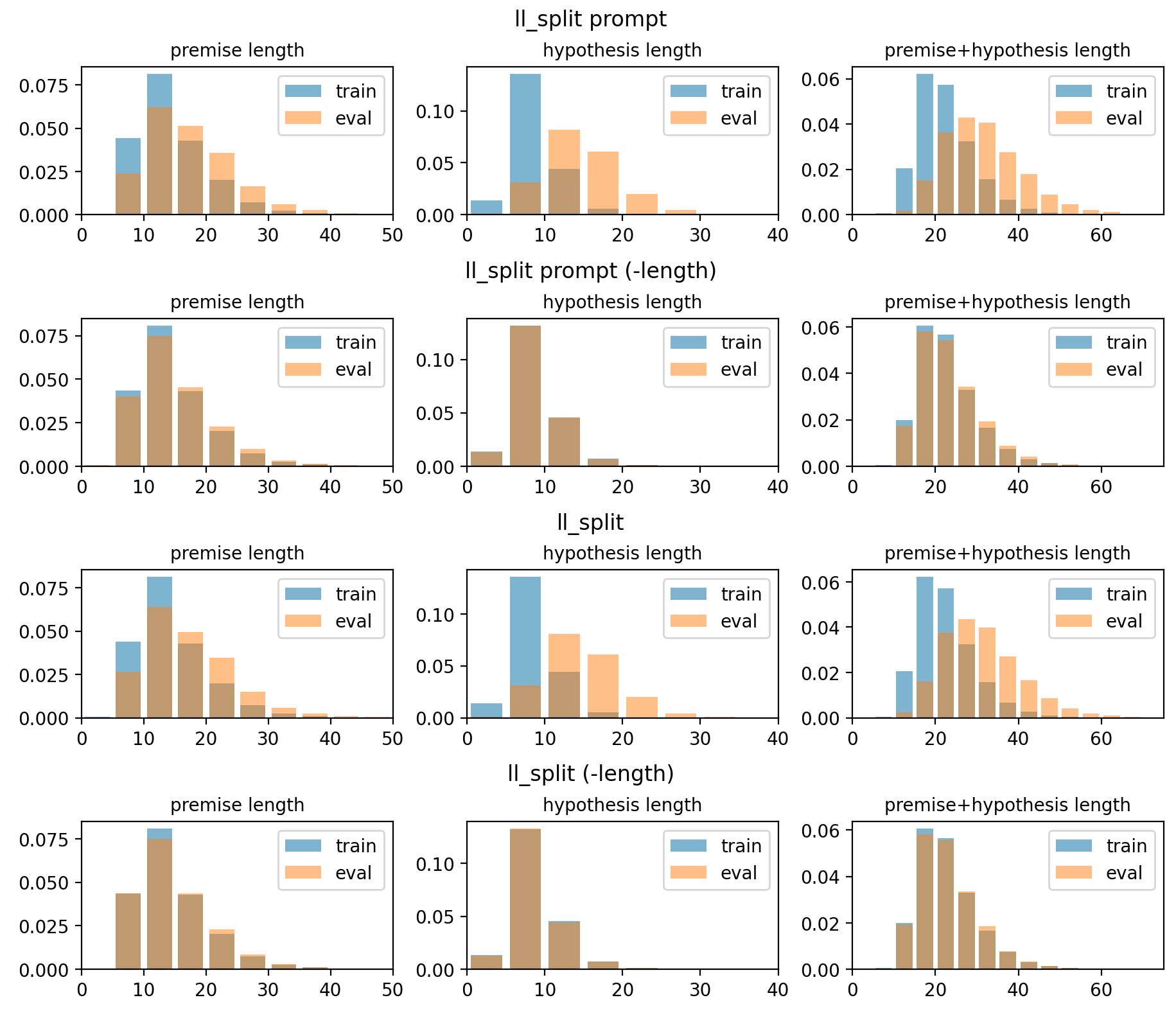}
\caption{\snli: Input length variation of premise and hypothesis for the splits.}
\label{fig:snli-inp-len-var}
\end{figure*}

From Figure~\ref{fig:snli-inp-len-var}, we see that the Likelihood Splits put longer premises and hypotheses in the evaluation set. Controlling for length completely removes this skew while increasing the difficulty of the splits (Table~\ref{tab:snli-acc}). This means that if we remove the factor of length from likelihood, the remaining examples have lower likelihood for other reasons; reasons that contribute to difficulty.

\subsection{\snli: Distribution of Rare Words}
\label{app:snli-rare-test-words-variation}

We report the fraction of test sets words that are rare for various splits in Figure~\ref{fig:snli-rare-test-words}. This evaluation combines the premise and the hypothesis i.e. it considers the full task input. In order to remove typographical errors, we only consider words that occur in a wordlist of English words (\url{https://github.com/dwyl/english-words}). We define rare words as words that occur at most 1 time per million words statistics collected in SUBTLEXus \citep{599801-subtlexus}. This process results in a list of 13478 low frequency words that occur in the \snli dataset. We find that Likelihood Splits put examples with a significantly large fraction or rare words in the evaluation set. Controlling for length increases the fraction of rare words since length is removed as a factor from likelihood.

\subsection{\snli: Variation of Syntactic complexity}
\label{app:snli-parse-variation}

\begin{figure*}[htb]
\centering
\includegraphics[width=0.8\textwidth]{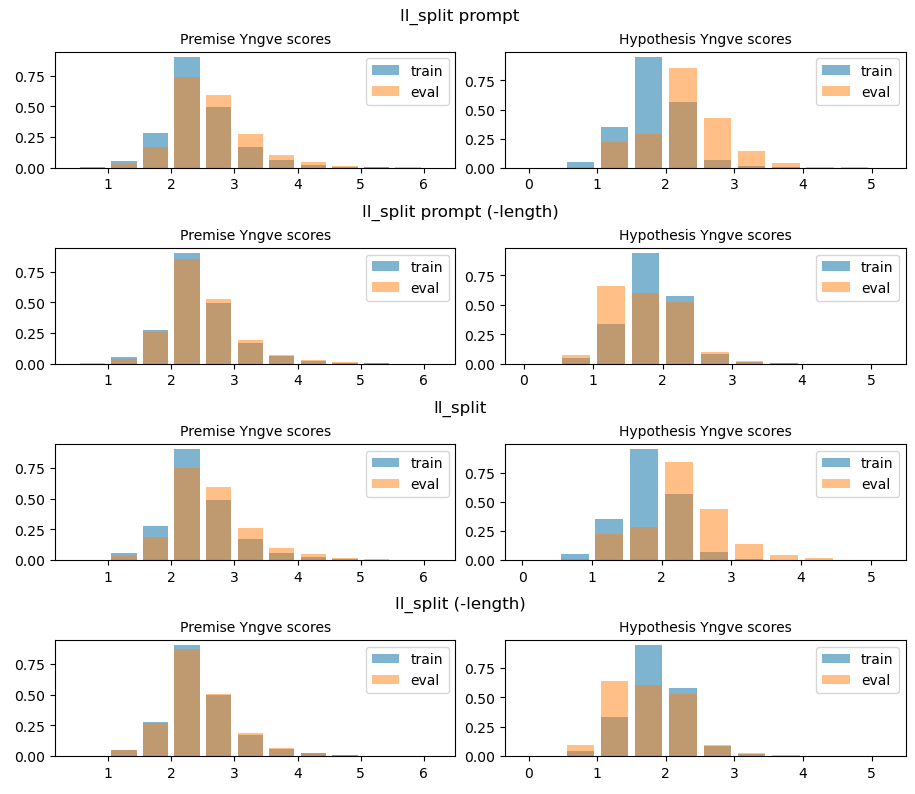}
\caption{\snli: Distribution of Yngve scores (syntactic complexity) computed on the parse tree of the premise and hypothesis. The dev sets for Likelihood Splits contain more complex utterances.}
\label{fig:snli-query-parse-stats}
\end{figure*}

We compute Yngve scores for premise and hypothesis of examples as described in Appendix~\ref{app:spider-parse-variation}. The complexity of premise and hypothesis in developments sets of Likelihood Splits is higher than in the corresponding train sets (see Figure~\ref{fig:snli-query-parse-stats}). Controlling for length removes this skew in the premise. However, the length controlled splits tend to have less syntactically complex hypotheses in the evaluation sets. This is surprising because the length-controlled variants are actually more difficult for the model; human performance is higher on length-controlled splits.

\subsection{\snli: Variation of Reading Level}
\label{app:snli-reading-variation}

\begin{figure*}[htb]
\centering
\includegraphics[width=0.9\textwidth]{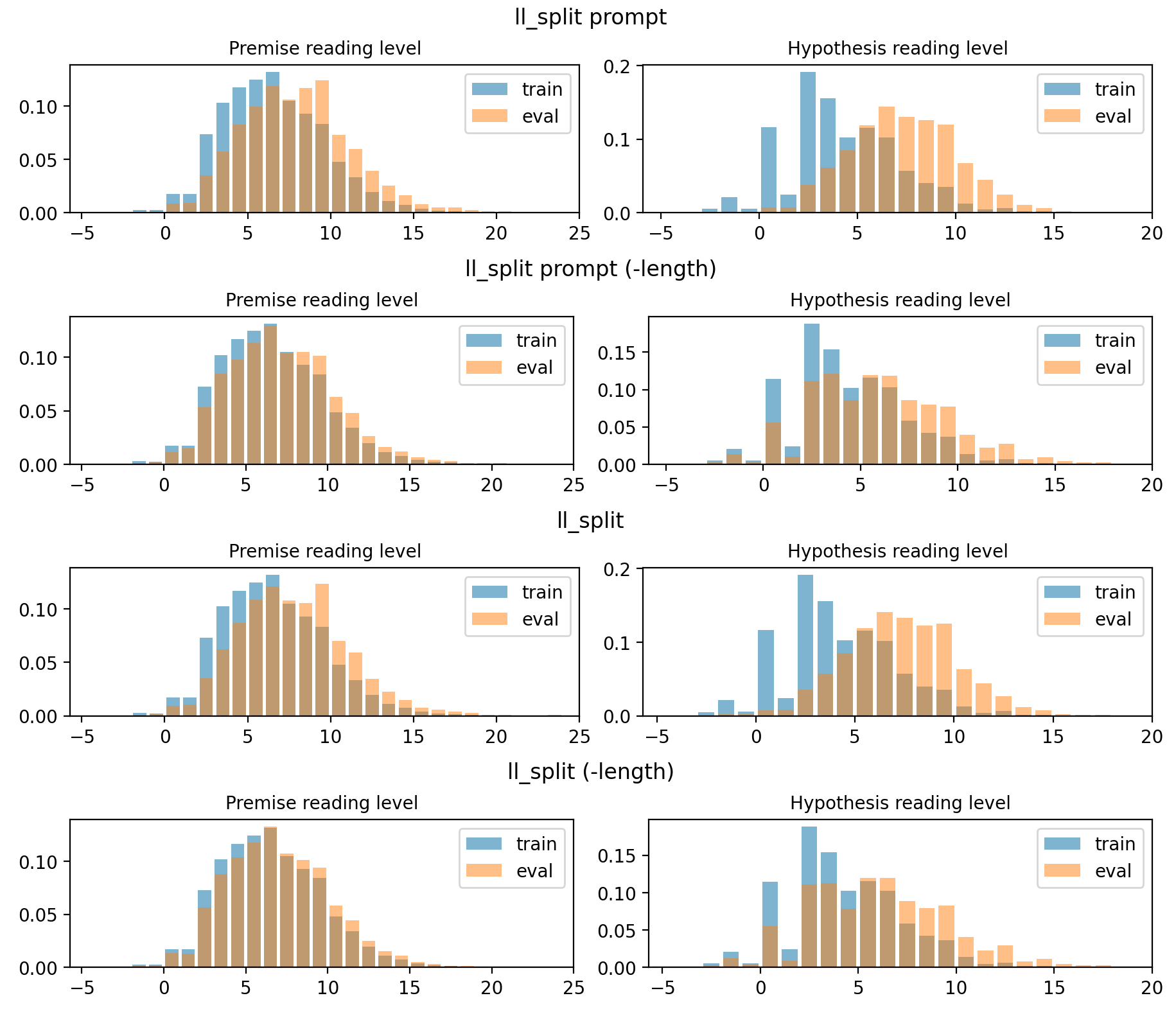}
\caption{\snli: Distribution of Flesch-Kincaid reading level for the premise and hypothesis in various splits. The dev sets for \emph{ll\_split} seem to contain more complex utterances. Normalizing for the length seems to reduce but not remove the effect.}
\label{fig:snli-reading-stats}
\end{figure*}

We compute the Flesch-Kincaid reading level \cite{Kincaid1975DerivationON} for premise and hypothesis of examples. This score takes into account the number of syllables per word in the sentence. The reading grade (complexity) of premise and hypothesis in developments sets of Likelihood Splits is higher than in the corresponding train sets (see Figure~\ref{fig:snli-reading-stats}). Controlling for length does not fully remove this skew and the evaluation examples retain more complex sentences than in the training set.

\subsection{\snli: Error Analysis}
\label{app:snli-err-analysis}

In Table~\ref{tab:snli-err-cases}, we present some examples from the development set of \emph{ll\_split (-len)} where the fine-tuned \roberta-large model predicts incorrectly. We divide them into categories: examples requiring external world knowledge, examples where a typo changes the meaning of the example, and examples with ambiguous or incorrect labels.

\begin{table*}[tb]
\centering
\footnotesize
\begin{tabular}{l p{4.5cm} p{3.5cm} c c}
\toprule
Error Type & Premise & Hypothesis & Gold Label & Predicted Label \\
\midrule[\heavyrulewidth]
\multirow{9}{2cm}{Requires External Knowledge} & A young boy is holding on and riding a zip line down a hill. & A exciting adventure! & entailment &  neutral \\
\cmidrule(lr){2-5}
& A young child is watching a toy construction brick construct. & A child is using lincoln logs. & neutral & contradiction \\
\cmidrule(lr){2-5}
& A performer is jumping off the stage into a crowd of fans. & The artist is crowdsurfing. & entailment & neutral \\
\cmidrule(lr){2-5}
& A couple holds up their child on a series of large steps while others are also traversing the steps. & A fourteen year old is restrained from the museum. & contradiction & neutral \\
\midrule
\multirow{2}{2cm}{Typo} & Martial artists perform in front of a crowd outdoors. & There is a crown outdoors. & entailment & neutral \\
\midrule
\multirow{4}{2cm}{Ambiguous / Incorrect Labels} & Technicians working in underground. & People work underground while dinosaurs attacked & neutral & contradiction \\
\cmidrule(lr){2-5}
& A young gentlemen with a blue tie talking into a microphone. & High winds will interfere with microphone recording. & entailment & neutral \\
\bottomrule
\end{tabular}
\caption{\label{tab:snli-err-cases}
\snli: Error analysis of \roberta-large on examples from \emph{ll\_split (-len)}.}
\end{table*}

\end{document}